\documentclass{article}
\usepackage[final]{colm2026_conference}

\usepackage{microtype}
\usepackage{hyperref}
\usepackage{url}
\usepackage{booktabs}
\usepackage{graphicx}
\usepackage{amsmath}
\usepackage{amssymb}
\usepackage{lineno}
\usepackage{multirow}
\usepackage{xcolor}

\definecolor{darkblue}{rgb}{0, 0, 0.5}
\hypersetup{colorlinks=true, citecolor=darkblue, linkcolor=darkblue, urlcolor=darkblue}

\title{Rule vs.\ Consequence: Dissociable Internal Representations of Moral Reasoning in LLMs}

\author{Eugenie Shi \\
Department of Computer Science\\
Stanford University\\
Stanford, CA 94305, USA \\
\texttt{\{yqshi\}@stanford.edu} \\
}

\begin{document}

\ifcolmsubmission
\linenumbers
\fi

\maketitle

\begin{abstract}
Large language models are increasingly deployed in morally and legally sensitive
contexts, yet little is known about how they internally represent distinctions
between moral reasoning styles. We ask whether a model's differential response to
rule-framed and consequence-framed moral questions reflects distinct internal
structure or merely surface lexical sensitivity. We construct a 2$\times$2 factorial
dataset crossing reasoning style (rule-based vs.\ consequence-based framing) with
judgment polarity (permissive vs.\ prohibitive), drawn from three established moral
reasoning benchmarks and evaluated on a cleanroom test set of novel phrasings that
includes conflict pairs where the two framings predict opposite labels. Using linear
probing and activation steering, we identify two orthogonal directions in the
residual stream: \textsc{Dir-A} tracks reasoning style and \textsc{Dir-B} tracks
judgment polarity, each near-blind to the other factor (cross-correlation
$|r| < 0.04$). This double dissociation replicates across three instruction-tuned
models from different developers at comparable proportional depth, and it is not
knife-edged in layer: the directions are recoverable, and causally effective,
throughout an extended band of the network. \textsc{Dir-A} is causally sufficient to
shift judgments (Cohen's $d > 1.0$), and on conflict scenarios steering selectively
corrects rule-framed errors while leaving consequence-framed judgments unchanged, a
pattern inconsistent with a global acceptance bias. Generalization beyond the
training templates is partial: the direction transfers across benchmark sources, but
its advantage over lexical baselines on novel phrasings is modest. We therefore frame
the result as moral reasoning style sensitivity, a geometrically dissociable
and causally effective structure, rather than internalized normative reasoning.
\end{abstract}

\section{Introduction}

As large language models are deployed in legally sensitive, medically critical, and ethically charged contexts \citep{soffer2025pitfalls, scherer2025normative}, understanding how they represent and respond to different moral framings becomes practically important. A fundamental distinction in moral philosophy separates rule-based reasoning, which judges actions by their conformity to explicit norms and duties \citep{rawls1971theory}, from consequence-based reasoning, which judges actions by their outcomes and welfare effects \citep{mill1863utilitarianism}.\footnote{We use \emph{rule-based} and \emph{consequence-based} rather than \emph{deontological} and \emph{utilitarian} deliberately. The philosophical terms name normative theories that carry commitments our stimuli do not probe, such as universalizability or the aggregation of welfare across persons, and our directions license no claims about those commitments. What we manipulate is how a question is posed. A model whose activations separate our two conditions has distinguished two ways of being asked, not adopted a moral theory. We reserve the theory names for the theories and use framing terms for the manipulation.} Models have been shown to produce systematically different judgments depending on how questions are framed \citep{simmons-2023-moral, park-etal-2025-deontological}, but whether this reflects superficial lexical sensitivity or distinct internal representational structures remains an open question with direct implications for alignment and interpretability research.

Mechanistic interpretability has established that LLMs encode semantic properties as linear directions in activation space \citep{park2023linear, burns2022discovering, zou2023representation, elhage2022superposition}, and that these directions can be causally manipulated via activation steering \citep{turner2023activation, panickssery2024steering}. We extend this program to moral reasoning style, asking whether the model's differential response to rule-framed and consequence-framed moral inputs has a geometric structure that is causally effective, dissociable from judgment polarity, and stable across diverse inputs and models.

To support this investigation, we construct a 2$\times$2 factorial dataset that crosses reasoning style with judgment polarity. We draw scenarios from ETHICS \citep{hendrycks2021aligning}, Moral Stories \citep{emelin2021moral}, and Social Chemistry 101 \citep{forbes2020social}, pairing each scenario with both a rule-framed and a consequence-framed question under matched polarity conditions. We additionally hand-write a cleanroom test set including conflict pairs where the two framings predict opposite gold labels, enabling behavioral dissociation analysis that goes beyond simple framing bias.

Using this dataset, we make four central contributions. First, we identify a double dissociation in which \textsc{Dir-A} encodes moral reasoning style ($r > 0.87$ for style, $r < 0.03$ for polarity) and \textsc{Dir-B} encodes judgment polarity ($r > 0.85$ for polarity, $r < 0.06$ for style), with cross-correlation $|r| < 0.04$, consistent across three models from different developers. Second, we demonstrate layer robustness: all ten layers tested in the range 6 to 15 of Qwen show significant framing selectivity ($p < 0.0001$), indicating that the geometric structure is a stable property of an entire processing region rather than a narrow artifact. Third, we provide causal evidence via activation steering showing large logit difference shifts (Cohen's $d > 1.0$, all $p < 0.0001$) and a framing-specific behavioral dissociation on conflict pairs that rules out a global acceptance-bias explanation. Fourth, we provide generalization evidence through cross-domain probing that transfers across three of four benchmark sources with 0.967 mean accuracy, and through lexical baseline comparisons showing the activation probe exceeds $n$-gram baselines on novel-template evaluation, though not a hand-specified keyword detector.

Throughout, we are careful to distinguish between our operational definition of \emph{moral reasoning style sensitivity}, meaning the model's differential response to rule-framed vs.\ consequence-framed inputs, and the stronger claim of fully internalized normative reasoning style. We argue the former is well supported by the evidence; the latter remains an open question.

\section{Related Work}

\paragraph{Linear representations in LLMs.}
\citet{park2023linear} demonstrate that semantic properties are encoded as linear directions in transformer residual streams. \citet{zou2023representation} show these directions are causally manipulable via representation engineering. \citet{burns2022discovering} identify truth-related directions through contrastive probing. \citet{elhage2022superposition} provide theoretical grounding for redundant encoding in superposition. Our work applies these methods to moral reasoning style.

\paragraph{Moral and normative reasoning in LLMs.}
Prior work evaluates LLM moral reasoning through behavioral benchmarks \citep{hendrycks2021aligning, emelin2021moral, forbes2020social}, finding systematic framing effects \citep{simmons-2023-moral}. \citet{park-etal-2025-deontological} show that deontological lexical markers and modal expressions systematically influence normative judgments. \citet{ozeki2025normative} benchmark normative reasoning from logical and modal perspectives. \citet{abrams2026norms} evaluate LLMs on norm-application tasks. \citet{zhong2025conformity} show that different judgment processes may be dissociable inside LLMs. \citet{soffer2025pitfalls} document failure modes in medical ethics reasoning. Our work moves beyond behavioral evaluation to examine the internal representational structure underlying these judgment patterns.

\paragraph{Framing effects, cognitive bias, and instruction sensitivity.}
\citet{itzhak-etal-2024-instructed} analyze emergent cognitive biases in instruction-tuned LLMs, including framing effects analogous to those we study, and link them to instruction phrasing. \citet{sun2024evaluating} demonstrate that instruction-tuned models are brittle to phrasing variations, motivating internal representation analysis rather than surface behavioral testing alone. \citet{lore2024strategic} show that LLM decisions in strategic games depend on contextual framing, providing complementary behavioral evidence. \citet{geng2026pragmatic} propose a framework to quantify pragmatic influences of instructions on LLM behavior. \citet{lim2026deframe} introduce DeFrame, a debiasing method that targets framing-induced biases; the directions we identify could serve as building blocks for such interventions. \citet{ceballos-arroyo-etal-2024-open} document instruction phrasing sensitivity in clinical LLMs, motivating the ethics statement of this work.

\paragraph{Activation steering and mechanistic analysis.}
\citet{turner2023activation} introduce activation addition and \citet{panickssery2024steering} extend it to personality traits. \citet{wu2025theory} probe internal representations of theory-of-mind concepts, providing a methodological comparison for our representational analysis. We apply activation steering to moral reasoning style, introducing a 2$\times$2 factorial design to disentangle reasoning style from judgment polarity.

\section{Dataset}

\subsection{Factorial design}

We construct a 2$\times$2 factorial dataset that crosses \textbf{moral reasoning style} (rule-based vs.\ consequence-based) with \textbf{judgment polarity} (permissive vs.\ prohibitive). Each example consists of a moral scenario paired with two questions: one that frames the judgment in terms of an applicable rule or norm, and one that frames it in terms of outcomes and welfare effects. The rule-framed template follows ``Given the social rule that [norm], is this action acceptable?'' and the consequence-framed template follows ``Considering the outcomes and effects on those involved, is this action acceptable?'' This design is essential for our analysis: by balancing reasoning style against polarity across the dataset, we create the conditions under which two dissociable directions can be identified rather than a single normative-strictness axis.

\subsection{Sources and splits}

We draw scenarios from three established moral reasoning benchmarks. ETHICS \citep{hendrycks2021aligning} provides scenarios from its deontology and commonsense subsets covering everyday moral violations and rule-following dilemmas. Moral Stories \citep{emelin2021moral} provides structured narratives about social norm violations. Social Chemistry 101 \citep{forbes2020social} provides social rules and judgments about everyday situations. Each scenario is paired with both a rule-framed and a consequence-framed question, and gold labels are assigned by the source benchmark. The resulting dataset contains 1,452 examples. The core moral split comprises 720 training, 240 validation, and 240 benchmark test examples, with strict label balance in the margins: 180 examples per reasoning style and polarity cell in the training set, and 60 per cell in each of validation and benchmark test. Two further splits are held out: a 32-item cleanroom test set of hand-written novel phrasings, and a 220-item legal transfer set drawn from LegalBench \citep{guha2023legalbench}.

We additionally construct a \textbf{cleanroom test set} of 32 hand-written items, 16 moral and 16 legal in subject matter, of which 20 form 10 conflict pairs in which rule-framed and consequence-framed questions on a shared scenario carry opposite gold labels. All phrasings are original to this set and absent from the training sources, enabling evaluation of generalization beyond learned templates.

\section{Methods}

\subsection{Models and activation extraction}

We use three instruction-tuned LLMs spanning different developers and tokenizers: Qwen2.5-7B-Instruct \citep{qwen2025} from Alibaba ($L=28$ layers, $d=3584$), Mistral-7B-Instruct-v0.3 \citep{jiang2023mistral} from Mistral AI ($L=32$, $d=4096$), and Llama-3.1-8B-Instruct \citep{dubey2024llama} from Meta ($L=32$, $d=4096$). All models are loaded in 4-bit NF4 quantization \citep{dettmers2023qlora} to fit within available GPU memory. Activations are extracted at the final prompt token from each transformer layer's residual stream, yielding a tensor of shape $[n, L, d]$ per split.

\subsection{Direction extraction}
\label{sec:directions}

For each layer $\ell$, we train $\ell_2$-regularized logistic regression probes ($C=1.0$) on standardized activations to separate rule-framed from consequence-framed examples (\textsc{Dir-A}) and permissive from prohibitive examples (\textsc{Dir-B}). The probe weight vector $\mathbf{w}$ is back-projected to the original activation space by $\mathbf{v} = \mathbf{w} / \boldsymbol{\sigma}$, where $\boldsymbol{\sigma}$ is the feature-wise standard deviation from the standardizer. Only the weight vector is used; the bias term encodes class prevalence rather than direction and is discarded. The resulting vector is $\ell_2$-normalized to give the unit-norm direction $\hat{\mathbf{v}}$. Layer selection uses the Fisher discriminant ratio $J = (\mu_+ - \mu_-)^2 / (\sigma_+^2 + \sigma_-^2)$ computed on held-out validation activations, where $\mu_\pm$ and $\sigma_\pm^2$ are the mean and variance of $\hat{\mathbf{v}}^\top \mathbf{x}$ for positive and negative class examples.

\subsection{Activation steering}

We inject directions into the residual stream at the final prompt token position via forward hooks:
\[
\mathbf{h}_\ell \leftarrow \mathbf{h}_\ell + \alpha\,\hat{\mathbf{v}}
\]
where $\hat{\mathbf{v}}$ is the unit-norm direction and $\alpha$ is the steering
strength. The absolute value of $\alpha$ is not interpretable in isolation; the
meaningful quantity is the \emph{perturbation ratio}, the steering vector norm
divided by the mean residual stream norm at the intervention layer. Four-bit NF4
quantization \citep{dettmers2023qlora} inflates absolute residual norms uniformly
(approximately 430 at layer 10 in Qwen, against approximately 50 in float16 models)
without changing the relative scale of an intervention. Our $\alpha = 40$ to $60$ for
Qwen therefore corresponds to perturbation ratios of $0.09$ to $0.14$, matching
$\alpha \approx 7$ in an unquantized model of the same architecture. We verify that
the intervention remains in-distribution: at $\alpha = 60$, $94.2$ percent of steered
residual dimensions fall within $[\mu - 3\sigma, \mu + 3\sigma]$ of the unsteered
activation distribution at layer 10. For Llama, $\alpha$ is calibrated to the same
perturbation ratio, yielding $\alpha = 2$; the fact that a calibrated $\alpha = 2$ in
one model and $\alpha = 60$ in another produce the same qualitative effects is itself
evidence that the ratio, not the raw magnitude, is the operative variable. We further
verify that steering effects are monotonically increasing within $|\alpha| \leq 60$
and report $\alpha = 60$ unless stated otherwise. Full details are in
Appendix~\ref{app:layer}.

\subsection{Measurement}

Model judgments are measured via logit difference:
\[
\Delta\text{LD} = \log\!\sum_{t \in \mathcal{A}} p_t \;-\; \log\!\sum_{t \in \mathcal{U}} p_t
\]
where $p_t = \text{softmax}(\mathbf{z})_t$ is the full-vocabulary softmax probability,
\[
\mathcal{A} = \{\text{``Acceptable''}, \text{``acceptable''}, \text{``Yes''}, \text{``yes''}\}
\]
and
\[
\mathcal{U} = \{\text{``Unacceptable''}, \text{``unacceptable''}, \text{``No''}, \text{``no''}\}
\]
This measure is robust to the near-binary output distribution observed in our data: 59 percent of benchmark examples have $P(\text{Acceptable}) < 0.1$, making probability-space shifts negligible while logit-space shifts remain meaningful. Note that this measurement operates only over these eight tokens and never over option letters. Where a source benchmark distributes items in multiple-choice form, we reformat the scenario into our binary acceptability frame and discard the original option labels, so no A/B/C/D token enters the logit difference at any point. Appendix~\ref{app:dataset} shows a source item alongside both of its templated forms.

\section{Results}

\subsection{Double dissociation across three models}
\label{sec:dissociation}

\begin{figure}[t]
\begin{center}
\includegraphics[width=0.95\linewidth]{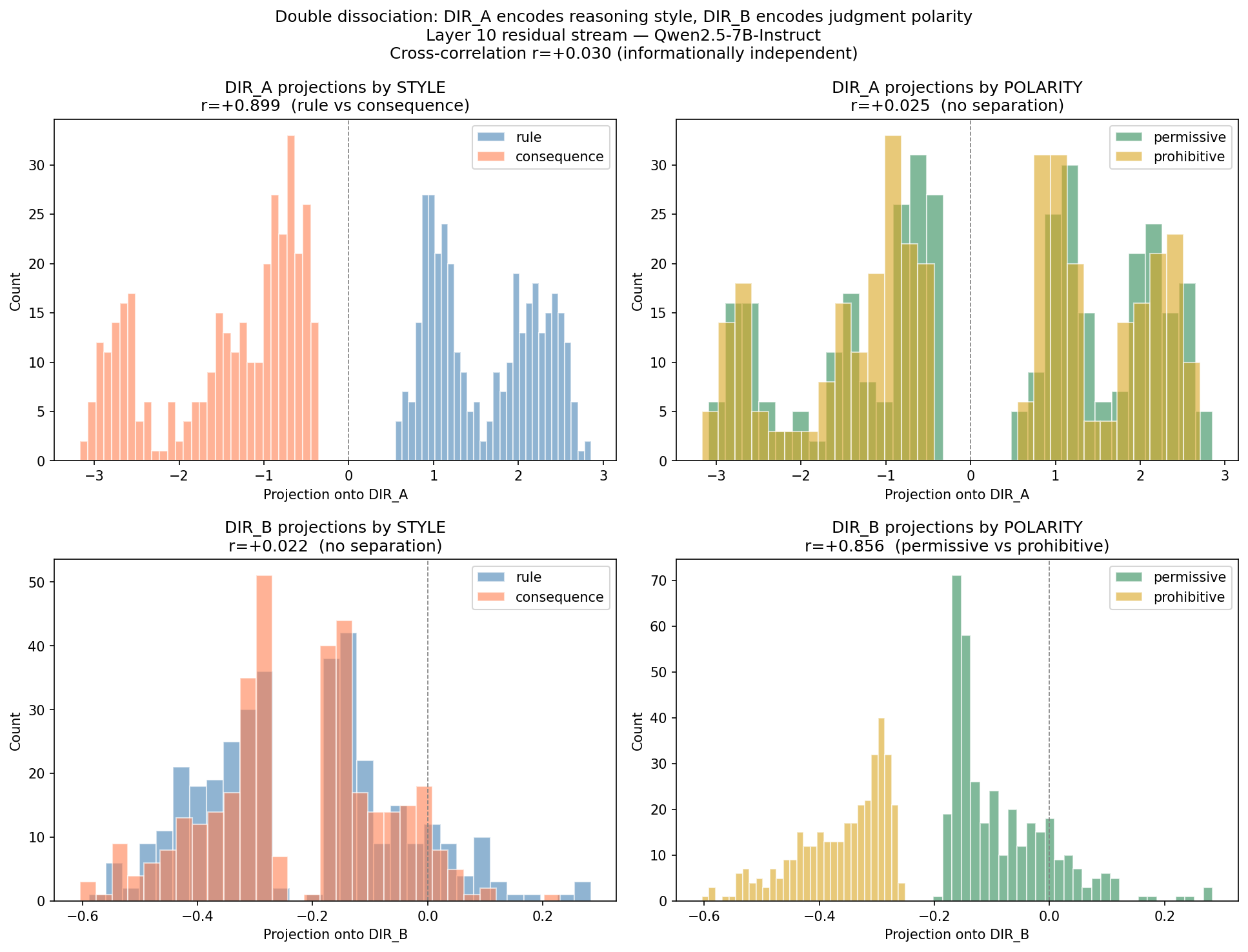}
\end{center}
\caption{Double dissociation across three LLMs. All models show \textsc{Dir-A} $|r(\text{style})| > 0.87$ and \textsc{Dir-B} $|r(\text{polarity})| > 0.85$, with cross-correlation $|r| < 0.04$.}
\label{fig:double_dissociation}
\end{figure}

Figure~\ref{fig:double_dissociation} shows the projections of all training examples onto \textsc{Dir-A} and \textsc{Dir-B} at layer 10 of Qwen. \textsc{Dir-A} cleanly separates rule-framed from consequence-framed examples ($r = +0.899$, $p < 0.0001$) with near-zero polarity correlation ($r = +0.025$, $p = 0.50$). \textsc{Dir-B} shows the complementary pattern: $r(\text{polarity}) = +0.856$ ($p < 0.0001$) and $r(\text{style}) = +0.022$ ($p = 0.56$). The cross-correlation between projections is $r = +0.030$ ($p = 0.43$), confirming informational independence. The complementary structure across all four panels of Figure~\ref{fig:double_dissociation} supports the two-direction interpretation over a single strictness axis.

\begin{figure}[t]
\begin{center}
\includegraphics[width=0.98\linewidth]{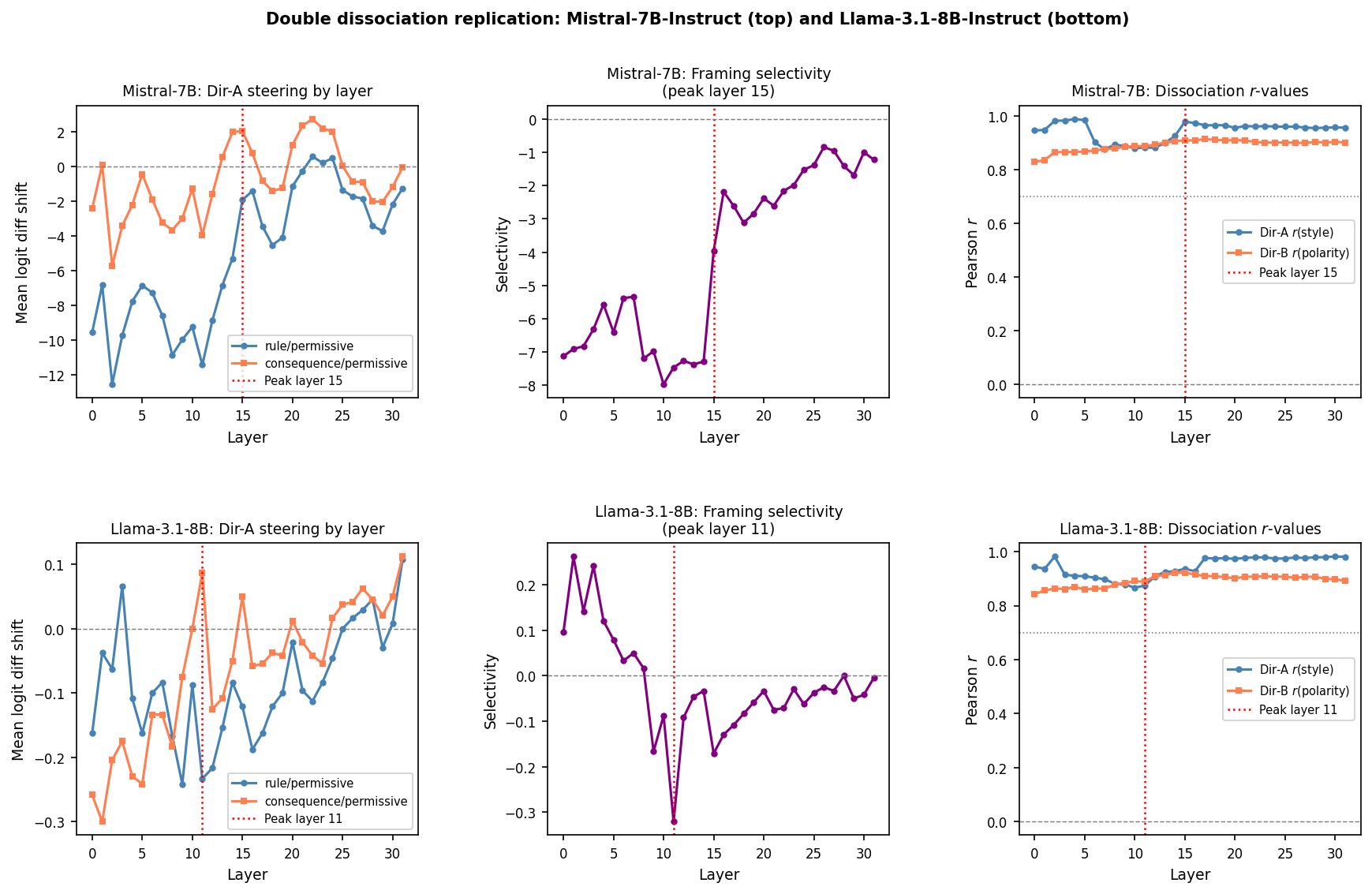}
\end{center}
\caption{Double dissociation replication in Mistral-7B-Instruct-v0.3 (top row)
and Llama-3.1-8B-Instruct (bottom row).}
\label{fig:replication}
\end{figure}

Table~\ref{tab:replication} reports the double dissociation across all three models. The pattern is consistent: \textsc{Dir-A} $|r(\text{style})| > 0.87$ and \textsc{Dir-B} $|r(\text{polarity})| > 0.85$ in every model, with cross-correlation $|r| < 0.04$. Peak dissociation layers fall at 34 to 47 percent network depth across models. This is consistent with the layer profiles in Section~\ref{sec:layerrobust} and Figure~\ref{fig:replication}, where framing selectivity is strongest at early-to-middle depth before decaying in later layers.

On the benchmark test set, the style probe achieves $1.000$ accuracy and
Dir-A projections separate style labels with AUC $= 1.000$; the polarity probe
likewise achieves $1.000$ accuracy on benchmark test. These perfect in-domain
scores confirm near-perfect linear separability at layer 10 and motivate
the cleanroom evaluation as the more stringent test of generalization.

\begin{table}[t]
\begin{center}
\begin{tabular}{lrrrrr}
\toprule
Model & Peak layer & \textsc{Dir-A} $r(\text{sty})$ & \textsc{Dir-A} $r(\text{pol})$ & \textsc{Dir-B} $r(\text{pol})$ & Cross-$r$ \\
\midrule
Qwen2.5-7B   & 10/28 (35.7\%) & $+0.899$ & $+0.025$ & $+0.856$ & $+0.030$ \\
Llama-3.1-8B & 11/32 (34.4\%) & $+0.875$ & $+0.002$ & $+0.888$ & $-0.032$ \\
Mistral-7B   & 15/32 (46.9\%) & $+0.978$ & $+0.007$ & $+0.909$ & $-0.021$ \\
\bottomrule
\end{tabular}
\end{center}
\caption{Double dissociation across three instruction-tuned LLMs from different developers.}
\label{tab:replication}
\end{table}

\subsection{Layer robustness}
\label{sec:layerrobust}

Two questions about layer choice are distinct, and we answer them separately: whether
the geometry exists only at one depth, and whether the causal effect survives moving the intervention layer.

On the first, Figure~\ref{fig:replication} reports the dissociation at every layer of
Mistral and Llama (32 each). \textsc{Dir-A} $r(\text{style})$ and \textsc{Dir-B}
$r(\text{polarity})$ exceed 0.85 across virtually all layers in both models, so the
directions are recoverable throughout the network rather than at a single privileged
depth.

On the second, we steer with the fixed layer-10 direction at each of layers 6 to 15 of
Qwen. We restrict the causal sweep to this band rather than the full 28 layers because
layer selection is governed by the Fisher discriminant ratio, which peaks sharply at
layer 10 ($J = 5.58$) and decays below $1.0$ by layer 15 (Appendix~\ref{app:layer}).
Intervening where the discriminant has decayed would test whether a direction carrying
little class signal still moves behavior, which is not a robustness check on the
present result. The informative question is whether the effect is knife-edged at the
discriminant peak, and Table~\ref{tab:layerrobust} shows that it is not: all ten
tested layers achieve significant framing selectivity ($p < 0.0001$), with the
qualitative pattern consistent throughout. Framing selectivity is the difference in
mean logit shift between rule-permissive and consequence-permissive examples under
\textsc{Dir-A}($+60$); negative selectivity is the expected sign, since
\textsc{Dir-A}($+60$) steers toward the consequence direction. The effect is a
property of a ten-layer processing region, not of a single hand-picked intervention
point. The corresponding visualization is Figure~\ref{fig:layerrobust} in
Appendix~\ref{app:layer}.

\subsection{Cross-domain generalization and lexical baselines}
\label{sec:generalization}

\paragraph{Cross-domain generalization.}

\begin{table}[t]
\begin{center}
\begin{tabular}{lrrr}
\toprule
Held-out source & $n_\text{test}$ & Accuracy & $\cos(\hat{v}_\text{cd}, \textsc{Dir-A})$ \\
\midrule
Social Chemistry 101 & 60 & 1.000 & 0.731 \\
Moral Stories        & 60 & 0.900 & 0.900 \\
ETHICS Commonsense   & 60 & 1.000 & 0.777 \\
ETHICS Deontology    & 60 & 0.517 & 0.835 \\
\midrule
Mean (excl.\ deontology) & & \textbf{0.967} & 0.803 \\
\bottomrule
\end{tabular}
\end{center}
\caption{Cross-domain generalization of \textsc{Dir-A} (Qwen, layer 10).}
\label{tab:crossdomain}
\end{table}

Table~\ref{tab:crossdomain} reports leave-one-source-out evaluation. Three key findings emerge:
\begin{itemize}
    \item Three of four sources achieve accuracy at or above 0.90 (mean 0.967 excluding the anomalous source).
    \item The ETHICS Deontology failure reflects its near-identical framing prefixes as both framings share the prefix ``Given that [excuse]...''. However, the direction maintains cosine similarity 0.835 to full-training \textsc{Dir-A}.
    \item Cross-domain cosine similarity to full-training \textsc{Dir-A} remains high across all sources (0.731--0.900), indicating the direction is geometrically stable even where the probe cannot exploit surface differences.
\end{itemize}

\paragraph{Lexical baselines.}
Table~\ref{tab:lexical} compares the activation probe against three lexical baselines. Both TF-IDF probes reach 1.000 in-domain but fall to chance on cleanroom (0.500, 0.438), showing their benchmark performance was attributable entirely to $n$-grams of the training templates. The activation probe retains 0.594, a 0.156 gain over the stronger variant, indicating it encodes information not reducible to those statistics. Keyword matching is different in kind: a hand-specified detector for framing markers rather than a fitted model. Because a question must still signal which framing it asks, some markers survive our cleanroom rewrites by construction, and keyword matching scores 0.750, above the activation probe. The cleanroom set defeats learned surface statistics but not a human-designed cue detector. We therefore claim only that the probe generalizes beyond the lexical statistics its training templates supply, not that it is the strongest available style classifier.

\begin{table}[t]
\begin{center}
\begin{tabular}{lrr}
\toprule
Probe type & Benchmark test & Cleanroom test \\
\midrule
Keyword matching          & 0.879 & 0.750 \\
TF-IDF (unigram)          & 1.000 & 0.500 \\
TF-IDF (bigram)           & 1.000 & 0.438 \\
Activation probe (layer 10) & 1.000 & \textbf{0.594} \\
\bottomrule
\end{tabular}
\end{center}
\caption{Probe accuracy on benchmark and cleanroom test sets.}\label{tab:lexical}
\end{table}

\subsection{Causal steering effects}
\label{sec:causal}

Table~\ref{tab:steering} reports mean logit difference shifts under \textsc{Dir-A} steering at $\alpha = \pm 60$ for each quadrant of the Qwen benchmark test set. All effects are statistically significant ($p \leq 0.003$) with large effect sizes. The \textsc{Dir-B} direction produces complementary effects: rule-permissive shift of $+1.113$ ($p < 0.0001$, Cohen's $d = 1.060$) and consequence-permissive shift of $-0.410$ ($p = 0.030$, $d = -0.290$), confirming that the two directions have qualitatively different behavioral signatures even within fixed-polarity conditions.

\begin{table}[t]
\begin{center}
\begin{tabular}{llrrrrr}
\toprule
Style & Polarity & $\alpha$ & Mean shift & Std & Cohen's $d$ & $p$ \\
\midrule
rule        & permissive  & $+60$ & $+2.025$ & $1.967$ & $+1.029$ & $<0.0001$ \\
rule        & prohibitive & $+60$ & $+2.877$ & $1.729$ & $+1.664$ & $<0.0001$ \\
consequence & permissive  & $-60$ & $-0.640$ & $1.605$ & $-0.398$ & $0.003$ \\
consequence & prohibitive & $-60$ & $-1.377$ & $1.212$ & $-1.137$ & $<0.0001$ \\
\bottomrule
\end{tabular}
\end{center}
\caption{Logit difference shifts under \textsc{Dir-A} steering ($n = 60$ per row, Qwen, $\alpha = 60$). All effects significant by paired $t$-test.}\label{tab:steering}
\end{table}

Steering with \textsc{Dir-A} improved overall accuracy from 0.613 to 0.650, a gain of 3.7 percentage points, with the largest absolute gain on consequence-permissive examples (13.3 percentage points, $0.183 \to 0.317$). This quadrant shows the lowest baseline accuracy, consistent with an RLHF-induced bias against consequence-framed permissive judgments that \textsc{Dir-A} partially counteracts.

At $|\alpha| = 40$ (monotonic regime), \textsc{Dir-A} produces mean $|\Delta\text{LD}| = 1.715$, compared to 0.778 and 0.640 for two random unit vectors, making \textsc{Dir-A} approximately 2.3 times more effective than random baselines and confirming that the direction captures semantically organized structure rather than generic perturbation effects.

Within fixed-polarity subsets, both directions produce significant effects: within permissive-only examples, \textsc{Dir-A}($+60$) shifts logit difference by $+2.242$ ($p < 0.0001$) and \textsc{Dir-B}($-60$) shifts it by $-1.108$ ($p < 0.0001$). This rules out a single normative-strictness-axis explanation in which style and polarity share a common representation.

Figure~\ref{fig:surgery} compares activation steering with prompt surgery,
in which the question template is rewritten from consequence-framed to
rule-framed at the text level. Steering along Dir-A and Dir-B produces logit
shifts of comparable magnitude to explicit prompt rewriting, providing
further evidence that the learned directions capture the rule-based and consequence-based framing signal
present in the input.

\subsection{Conflict pair behavioral dissociation}
\label{sec:conflict}

The cleanroom conflict pairs, where rule-framed and consequence-framed questions predict opposite gold labels, reveal a framing-specific behavioral pattern under \textsc{Dir-A} steering at $\alpha = +60$. For rule-framed conflict examples, where the baseline model achieves only 0.400 accuracy, steering produces a 30 percentage point improvement ($0.400 \to 0.700$, with 3 prediction flips). For consequence-framed conflict examples, where baseline accuracy is already 0.800, steering produces zero prediction flips despite a comparable mean logit shift of $+2.61$.

The behavioral asymmetry is the key finding: \textsc{Dir-A} selectively corrects errors in rule-framed conflict scenarios while leaving consequence-framed accuracy unchanged. This asymmetry cannot be explained by a global acceptance-direction account, which would predict proportional flipping in both framing conditions. The 0 out of 10 flip rate on consequence-framed examples, despite a large logit shift, is consistent with a ceiling effect: these examples are correctly classified with high confidence, so even large perturbations cannot change the discrete prediction. The result suggests framing-specific causal leverage: the direction interacts differently with rule-framed and consequence-framed representations even when applied with identical magnitude.

Using a Wilson score interval, the 95\% confidence interval for the
rule-framed post-steering accuracy is $[0.349, 0.910]$ (observed
$0.700$), and for consequence-framed accuracy is $[0.483, 0.957]$ (observed $0.800$). We acknowledge that with $n=10$ per condition these
intervals are wide, and treat this as behavioral evidence
consistent with the mechanistic account rather than a definitive statistical
claim.

\subsection{Approximate commutativity as evidence of independence}
\label{sec:commute}

We test order-dependence of sequential \textsc{Dir-A} followed by \textsc{Dir-B} versus the reverse order, and compare against a permutation-based null distribution of 100 random direction pairs applied at the same two layers (5 and 10). The Style-to-Polarity versus Polarity-to-Style order-dependence (mean $|\text{AB} - \text{BA}| = 0.719$) falls at the 8th percentile of the random-pair null distribution (mean $0.915$, standard deviation $0.134$, $t = -4.20$, $p < 0.0001$). The two directions are significantly more commutative than typical random direction pairs. This is consistent with functionally independent computational pathways: if the directions shared circuitry, their sequential application would show interference effects elevating order-dependence above the random baseline.

\subsection{Ablation}
\label{sec:ablation}

Orthogonal ablation of \textsc{Dir-A} and \textsc{Dir-B} at layer 10 produced near-zero selective effects, with selectivity ratios below 1.0. This indicates the directions are not uniquely necessary for the model's normative judgments, consistent with redundant encoding across multiple circuits \citep{elhage2022superposition}. Causal sufficiency is established via steering; necessity is not, and we do not claim it.

\section{Discussion}

\paragraph{What the results establish.}
The geometric orthogonality of \textsc{Dir-A} and \textsc{Dir-B} ($|r| < 0.04$ across three models) establishes that the model does not represent moral reasoning style and judgment polarity on a single strictness axis. The causal evidence (Cohen's $d > 1.0$, conflict pair asymmetry) establishes that \textsc{Dir-A} is upstream of behavior and not merely correlated with surface features. The three-model consistency across Qwen, Llama, and Mistral from three different developers, combined with consistent proportional depth (34 to 47 percent), establishes that this geometric organization is not a model-specific artifact. The layer robustness finding across layers 6 to 15 (all $p < 0.0001$) strengthens this further: the effect is achievable across an entire processing region, not just at one specific point.

\paragraph{Style sensitivity versus internalized reasoning.}
The identified directions reflect the model's differential response to rule-framed and consequence-framed moral inputs. Whether this sensitivity reflects genuine internalization of moral reasoning styles or learned lexical pattern matching cannot be fully determined. Three findings point toward representational structure beyond surface sensitivity: cross-domain generalization of 0.967 across three diverse sources, an activation probe advantage of 0.156 over the strongest $n$-gram baseline on novel templates, and framing-specific behavioral dissociation on conflict pairs. Against this, the probe does not exceed a hand-specified keyword detector on the cleanroom set (0.594 vs.\ 0.750), and the absence of correlation between \textsc{Dir-A} projection and spontaneous reasoning style on neutral prompts ($r = -0.056$, $p = 0.79$) means we cannot fully exclude surface-sensitivity explanations. We therefore adopt the conservative framing of moral reasoning style sensitivity throughout.

\paragraph{Error analysis.}
Among cleanroom test examples where the activation probe errs, two patterns dominate: examples where both framing conditions share similar opening phrases (consistent with the ETHICS Deontology finding), and examples where scenario content itself contains explicit normative language that partially overrides question framing. This suggests the direction is most reliable when question framing is the primary style-distinguishing signal rather than scenario content.

\paragraph{Rule-acceptance bias.}
\textsc{Dir-A}($+60$) increases logit difference across all four benchmark quadrants, reflecting a learned prior in the RLHF training distribution that rule-framing is associated with permissive judgments. The direction is nonetheless a genuine style-sensitivity direction, as confirmed by its $r = 0.899$ with style labels and $r = 0.025$ with polarity labels, and by the selective conflict pair effects.

\paragraph{Limitations.}
The large $\alpha$ values required (40 to 60) are an artifact of 4-bit quantization rather than an unusually aggressive intervention, as the perturbation ratio analysis in Appendix~\ref{app:layer} establishes; nonetheless, we have not replicated at full precision, and $|\alpha| > 60$ induces non-monotonic, likely out-of-distribution behavior. The cleanroom conflict set is relatively small. The factorial design is balanced in its margins but not within source (Appendix~\ref{app:dataset}), so reasoning style is partially predictable from source identity. A within-source crossed design would isolate the style manipulation more cleanly. Probe accuracy on cleanroom examples (0.594) indicates partial rather than full generalization across surface forms. All models tested are 7 to 8 billion parameter instruction-tuned chat variants, and base models and larger frontier models may show different geometric organization. The rule-acceptance bias observed in Section 6
suggests RLHF may amplify or create the style-sensitive direction, and
testing base variants of Llama and Mistral is an important direction
for future work. Future work should examine full-precision models, expand the conflict pair set, and test whether the direction predicts spontaneous reasoning style in chain-of-thought settings \citep{wang2022self}. Finally, we establish that reasoning style is dissociable and steerable, but not that steering it confers control on a downstream task of independent interest. The natural next step is to test whether \textsc{Dir-A} supports targeted control on an applied benchmark, either by holding judgment polarity fixed while shifting the framing a model reasons from, or by auditing whether a deployed model's normative outputs are framing-driven. We regard this as the most valuable extension of the present work.

\section{Conclusion}

We construct a 2$\times$2 factorial moral reasoning dataset crossing reasoning style with judgment polarity, drawing from ETHICS, Moral Stories, and Social Chemistry 101, and use it to identify two orthogonal directions in the residual streams of three instruction-tuned LLMs that separately encode moral reasoning style sensitivity (\textsc{Dir-A}) and judgment polarity (\textsc{Dir-B}). The double dissociation is consistent across Qwen2.5-7B, Llama-3.1-8B, and Mistral-7B, with $|r| > 0.85$ on target, $|r| < 0.06$ off-target, and $|r| < 0.04$ cross-correlation across all models. The framing-sensitive direction is robust across all ten layers tested in the range 6 to 15 (all $p < 0.0001$), indicating a stable geometric property of an entire processing region rather than a narrow artifact. \textsc{Dir-A} is causally effective (Cohen's $d > 1.0$), transfers across three of four benchmark domains (0.967 accuracy), and produces a framing-specific behavioral dissociation on conflict pairs that rules out a global acceptance-bias explanation. Sequential application of the two directions is significantly more commutative than random direction pairs, consistent with independent computational pathways. These results establish that instruction-tuned LLMs organize moral reasoning style sensitivity as a geometrically dissociable and causally effective structure, providing a foundation for mechanistic understanding of moral judgment in LLMs.

\section*{Ethics Statement}
This work analyzes internal representations of publicly available LLMs on publicly available moral reasoning benchmarks. No human subjects were involved. The ability to selectively shift model moral judgments via activation steering carries dual-use risks: such techniques could be used to bias outputs in legally sensitive, medical, or safety-related contexts toward a preferred framing without improving underlying reliability. We do not advocate deployment of steering interventions in normative systems without extensive safety evaluation \citep{scherer2025normative}. The findings reflect the specific annotation conventions and social assumptions embedded in the benchmark sources and the RLHF training pipelines of the evaluated models, and they should not be interpreted as universal or culturally robust normative structure \citep{ceballos-arroyo-etal-2024-open}.

\bibliography{colm2026_conference}

@book{rawls1971theory,
  author    = {Rawls, John},
  title     = {A Theory of Justice},
  publisher = {Harvard University Press},
  year      = {1971}
}

@book{mill1863utilitarianism,
  author    = {Mill, John Stuart},
  title     = {Utilitarianism},
  publisher = {Parker, Son, and Bourn},
  year      = {1863}
}

@inproceedings{park2023linear,
  author    = {Park, Kiho and Choe, Yo Joong and Veitch, Victor},
  title     = {The Linear Representation Hypothesis in Language Models},
  booktitle = {NeurIPS 2023 Workshop on Causal Representation Learning},
  year      = {2023},
  url       = {https://neurips.cc/virtual/2023/77537}
}

@inproceedings{burns2022discovering,
  author    = {Burns, Collin and Ye, Haotian and Klein, Dan and Steinhardt, Jacob},
  title     = {Discovering Latent Knowledge in Language Models Without Supervision},
  booktitle = {International Conference on Learning Representations},
  year      = {2023}
}

@article{zou2023representation,
  author    = {Zou, Andy and Phan, Long and Chen, Sarah and Campbell, James and Guo, Phillip and Ren, Richard and Pan, Alexander and Yin, Xuwang and Mazeika, Mantas and Dombrowski, Ann-Kathrin and Goel, Shashwat and Li, Nathaniel and Byun, Michael J. and Wang, Zifan and Mallen, Alex and Basart, Steven and Koyejo, Sanmi and Song, Dawn and Fredrikson, Matt and Kolter, J. Zico and Hendrycks, Dan},
  title     = {Representation Engineering: A Top-Down Approach to {AI} Transparency},
  journal   = {arXiv preprint arXiv:2310.01405},
  year      = {2023}
}

@article{elhage2022superposition,
  author    = {Elhage, Nelson and Hume, Tristan and Olsson, Catherine and Schiefer, Nicholas and others},
  title     = {Toy Models of Superposition},
  journal   = {Transformer Circuits Thread},
  year      = {2022},
  url       = {https://transformer-circuits.pub/2022/toy_model/index.html}
}

@article{turner2023activation,
  author    = {Turner, Alexander Matt and Thiergart, Lisa and
               Leech, Gavin and Udell, David and Vazquez, Juan J. and
               Mini, Ulisse and MacDiarmid, Monte},
  title     = {Steering Language Models With Activation Engineering},
  journal   = {arXiv preprint arXiv:2308.10248},
  year      = {2023},
  url       = {https://arxiv.org/abs/2308.10248}
}

@article{panickssery2024steering,
  author    = {Panickssery, Nina and Gabrieli, Nick and Schulz, Julian and Tong, Meg and Hubinger, Evan and Turner, Alexander Matt},
  title     = {Steering {LLaMA 2} via Contrastive Activation Addition},
  journal   = {arXiv preprint arXiv:2312.06681},
  year      = {2024}
}

@inproceedings{hendrycks2021aligning,
  author    = {Hendrycks, Dan and Burns, Collin and Basart, Steven and Critch, Andrew
               and Li, Jerry and Song, Dawn and Steinhardt, Jacob},
  title     = {Aligning {AI} With Shared Human Values},
  booktitle = {International Conference on Learning Representations},
  year      = {2021}
}

@inproceedings{emelin2021moral,
  author    = {Emelin, Denis and Le Bras, Ronan and Hwang, Jena D. and
               Forbes, Maxwell and Choi, Yejin},
  title     = {Moral Stories: Situated Reasoning about Norms, Intents,
               Actions, and their Consequences},
  booktitle = {Proceedings of the 2021 Conference on Empirical Methods
               in Natural Language Processing},
  pages     = {6905--6926},
  year      = {2021},
  url       = {https://aclanthology.org/2021.emnlp-main.54}
}

@inproceedings{forbes2020social,
  author    = {Forbes, Maxwell and Hwang, Jena D. and Shwartz, Vered
               and Sap, Maarten and Choi, Yejin},
  title     = {Social Chemistry 101: Learning to Reason about Social and Moral Norms},
  booktitle = {Proceedings of the 2020 Conference on Empirical Methods
               in Natural Language Processing},
  pages     = {653--670},
  year      = {2020},
  url       = {https://aclanthology.org/2020.emnlp-main.48}
}

@inproceedings{simmons-2023-moral,
    title = "Moral Mimicry: Large Language Models Produce Moral Rationalizations Tailored to Political Identity",
    author = "Simmons, Gabriel",
    editor = "Padmakumar, Vishakh  and
      Vallejo, Gisela  and
      Fu, Yao",
    booktitle = "Proceedings of the 61st Annual Meeting of the Association for Computational Linguistics (Volume 4: Student Research Workshop)",
    month = jul,
    year = "2023",
    address = "Toronto, Canada",
    publisher = "Association for Computational Linguistics",
    url = "https://aclanthology.org/2023.acl-srw.40/",
    doi = "10.18653/v1/2023.acl-srw.40",
    pages = "282--297"
}

@inproceedings{park-etal-2025-deontological,
    title = "Deontological Keyword Bias: The Impact of Modal Expressions on Normative Judgments of Language Models",
    author = "Park, Bumjin  and
      Lee, Jinsil  and
      Choi, Jaesik",
    editor = "Che, Wanxiang  and
      Nabende, Joyce  and
      Shutova, Ekaterina  and
      Pilehvar, Mohammad Taher",
    booktitle = "Proceedings of the 63rd Annual Meeting of the Association for Computational Linguistics (Volume 1: Long Papers)",
    month = jul,
    year = "2025",
    address = "Vienna, Austria",
    publisher = "Association for Computational Linguistics",
    url = "https://aclanthology.org/2025.acl-long.360/",
    doi = "10.18653/v1/2025.acl-long.360",
    pages = "7277--7296",
    ISBN = "979-8-89176-251-0"
}

@article{ozeki2025normative,
  author    = {Ozeki, Kentaro and Ando, Risako and Morishita, Takanobu and Abe, Hirohiko and Mineshima, Koji and Okada, Mitsuhiro},
  title     = {Normative Reasoning in Large Language Models: A Comparative Benchmark from Logical and Modal Perspectives},
  journal   = {arXiv preprint arXiv:2510.26606},
  year      = {2025}
}

@article{abrams2026norms,
  author    = {Abrams, Mitchell and Eskandari Miandoab, Kaveh and Gervits, Felix and Sarathy, Vasanth and Scheutz, Matthias},
  title     = {Where Norms and References Collide: Evaluating {LLMs} on Normative Reasoning},
  journal   = {arXiv preprint arXiv:2602.02975},
  year      = {2026}
}

@article{zhong2025conformity,
  author    = {Zhong, Huixin and Liu, Yanan and Cao, Qi and
               Wang, Shijin and Ye, Zijing and Wang, Zimu and
               Zhang, Shiyao},
  title     = {Disentangling the Drivers of {LLM} Social Conformity:
               An Uncertainty-Moderated Dual-Process Mechanism},
  journal   = {arXiv preprint arXiv:2508.14918},
  year      = {2025},
  url       = {https://arxiv.org/abs/2508.14918}
}

@article{soffer2025pitfalls,
  author    = {Soffer, Shelly and Sorin, Vera and Nadkarni, Girish N.
               and Klang, Eyal},
  title     = {Pitfalls of Large Language Models in Medical Ethics
               Reasoning},
  journal   = {npj Digital Medicine},
  volume    = {8},
  pages     = {461},
  year      = {2025},
  doi       = {10.1038/s41746-025-01792-y},
  url       = {https://www.nature.com/articles/s41746-025-01792-y}
}

@article{scherer2025normative,
  author    = {Millière, Raphaël 
},
  title     = {Normative Conflicts and Shallow {AI} Alignment},
  journal   = {arXiv preprint arXiv:2506.04679},
  year      = {2025}
}

@article{itzhak-etal-2024-instructed,
    title = "Instructed to Bias: Instruction-Tuned Language Models Exhibit Emergent Cognitive Bias",
    author = "Itzhak, Itay  and
      Stanovsky, Gabriel  and
      Rosenfeld, Nir  and
      Belinkov, Yonatan",
    journal = "Transactions of the Association for Computational Linguistics",
    volume = "12",
    year = "2024",
    address = "Cambridge, MA",
    publisher = "MIT Press",
    url = "https://aclanthology.org/2024.tacl-1.43/",
    doi = "10.1162/tacl_a_00673",
    pages = "771--785"
}

@article{sun2024evaluating,
  author    = {Sun, Tianhao and Shaib, Chaimaa and Wallace, Byron C.},
  title     = {Evaluating the Zero-Shot Robustness of Instruction-Tuned
               Language Models},
  journal   = {arXiv preprint arXiv:2306.11270},
  year      = {2024},
  url       = {https://arxiv.org/abs/2306.11270}
}

@article{lore2024strategic,
  author    = {Lor{\`{e}}, Nunzio and Heydari, Babak},
  title     = {Strategic Behavior of Large Language Models and the Role of Game Structure Versus Contextual Framing},
  journal   = {arXiv preprint arXiv:2309.05898},
  year      = {2024}
}

@article{geng2026pragmatic,
  author    = {Geng, Yilin and Abend, Omri and Hovy, Eduard and Frermann, Lea},
  title     = {Measuring Pragmatic Influence in Large Language Model Instructions},
  journal   = {arXiv preprint arXiv:2602.21223},
  year      = {2026}
}

@article{lim2026deframe,
  author    = {Lim, Kahee and Kim, Soyeon and Whang, Steven Euijong},
  title     = {{DeFrame}: Debiasing Large Language Models Against Framing Effects},
  journal   = {arXiv preprint arXiv:2602.04306},
  year      = {2026}
}

@inproceedings{ceballos-arroyo-etal-2024-open,
    title = "Open (Clinical) {LLM}s are Sensitive to Instruction Phrasings",
    author = "Ceballos-Arroyo, Alberto Mario  and
      Munnangi, Monica  and
      Sun, Jiuding  and
      Zhang, Karen  and
      McInerney, Jered  and
      Wallace, Byron C.  and
      Amir, Silvio",
    editor = "Demner-Fushman, Dina  and
      Ananiadou, Sophia  and
      Miwa, Makoto  and
      Roberts, Kirk  and
      Tsujii, Junichi",
    booktitle = "Proceedings of the 23rd Workshop on Biomedical Natural Language Processing",
    month = aug,
    year = "2024",
    address = "Bangkok, Thailand",
    publisher = "Association for Computational Linguistics",
    url = "https://aclanthology.org/2024.bionlp-1.5/",
    doi = "10.18653/v1/2024.bionlp-1.5",
    pages = "50--71"
}

@article{wu2025theory,
  author    = {Wu, Yuheng and Guo, Wentao and Liu, Zirui and
               Ji, Heng and Xu, Zhaozhuo and Zhang, Denghui},
  title     = {How Large Language Models Encode Theory-of-Mind:
               A Study on Sparse Parameter Patterns},
  journal   = {npj Artificial Intelligence},
  volume    = {1},
  pages     = {20},
  year      = {2025},
  doi       = {10.1038/s44387-025-00031-9},
  url       = {https://www.nature.com/articles/s44387-025-00031-9}
}

@article{qwen2025,
  author    = {Yang, An and Yang, Baosong and Zhang, Beichen and
               Hui, Binyuan and Zheng, Bo and Yu, Bowen and
               Li, Chengyuan and Liu, Dayiheng and Huang, Fei and
               Wei, Haoran and Lin, Huan and Yang, Jian and
               Tu, Jianhong and Zhang, Jianwei and Yang, Jianxin and
               Yang, Jiaxi and Zhou, Jingren and Lin, Junyang and
               Dang, Kai and Lu, Keming and Bao, Keqin and
               Yang, Kexin and Yu, Le and Li, Mei and Xue, Mingfeng and
               Zhang, Pei and Zhu, Qin and Men, Rui and Lin, Runji and
               Li, Tianhao and Tang, Tianyi and Xia, Tingyu and
               Ren, Xingzhang and Ren, Xuancheng and Fan, Yang and
               Su, Yang and Zhang, Yichang and Wan, Yu and
               Liu, Yuqiong and Cui, Zeyu and Zhang, Zhenru and
               Qiu, Zihan and others},
  title     = {{Qwen2.5} Technical Report},
  journal   = {arXiv preprint arXiv:2412.15115},
  year      = {2025},
  url       = {https://arxiv.org/abs/2412.15115}
}

@inproceedings{jiang2023mistral,
  author    = {Jiang, Albert Q. and Sablayrolles, Alexandre and Mensch, Arthur
               and Bamford, Chris and Chaplot, Devendra Singh and
               de las Casas, Diego and Bressand, Florian and Lengyel, Gianna
               and Lample, Guillaume and Saulnier, Lucile and
               Renard Lavaud, L\'{e}lio and Lachaux, Marie-Anne and Stock, Pierre
               and Le Scao, Teven and Lavril, Thibaut and Wang, Thomas and
               Lacroix, Timoth\'{e}e and El Sayed, William},
  title     = {Mistral {7B}},
  journal   = {arXiv preprint arXiv:2310.06825},
  year      = {2023}
}

@article{dubey2024llama,
  author    = {Dubey, Abhimanyu and Jauhri, Abhinav and Pandey, Abhinav and Kadian, Abhishek and Al-Dahle, Ahmad and Letman, Aiesha and others},
  title     = {The {Llama 3} Herd of Models},
  journal   = {arXiv preprint arXiv:2407.21783},
  year      = {2024}
}

@inproceedings{dettmers2023qlora,
  author    = {Dettmers, Tim and Pagnoni, Artidoro and Holtzman, Ari
               and Zettlemoyer, Luke},
  title     = {{QLoRA}: Efficient Finetuning of Quantized {LLMs}},
  booktitle = {Proceedings of the 37th International Conference on
               Neural Information Processing Systems},
  year      = {2023}
}

@inproceedings{wang2022self,
  author    = {Wang, Yizhong and Kordi, Yeganeh and Mishra, Swaroop and Liu, Alisa
               and Smith, Noah A. and Khashabi, Daniel and Hajishirzi, Hannaneh},
  title     = {Self-Instruct: Aligning Language Models with Self-Generated
               Instructions},
  booktitle = {Proceedings of the 61st Annual Meeting of the Association
               for Computational Linguistics},
  pages     = {13484--13508},
  year      = {2023},
  url       = {https://arxiv.org/abs/2212.10560}
}

@inproceedings{guha2023legalbench,
  author    = {Guha, Neel and Nyarko, Julian and Ho, Daniel E. and R\'{e}, Christopher
               and Chilton, Adam and Narayana, Aditya and Chohlas-Wood, Alex and
               Peters, Austin and Waldon, Brandon and Rockmore, Daniel N. and
               Zambrano, Diego and Talisman, Dmitry and Hoque, Enam and Surani, Faiz
               and Fagan, Frank and Sarfaty, Galit and Dickinson, Gregory M. and
               Porat, Haggai and Hegland, Jason and Wu, Jessica and Nudell, Joe and
               Niklaus, Joel and Nay, John and Choi, Jonathan H. and Tobia, Kevin and
               Hagan, Margaret and Ma, Megan and Livermore, Michael and
               Rasumov-Rahe, Nikon and Holzenberger, Nils and Kolt, Noam and
               Henderson, Peter and Rehaag, Sean and Goel, Sharad and Gao, Shang and
               Williams, Spencer and Gandhi, Sunny and Zur, Tom and Iyer, Varun and
               Li, Zehua},
  title     = {{LegalBench}: A Collaboratively Built Benchmark for Measuring
               Legal Reasoning in Large Language Models},
  booktitle = {Advances in Neural Information Processing Systems},
  year      = {2023}
}
\bibliographystyle{colm2026_conference}

\appendix

\section{Dataset construction details}
\label{app:dataset}

The 2$\times$2 factorial design pairs each source scenario with two question templates. Rule-framed questions follow the template ``Given [applicable norm], is this action acceptable?'' Consequence-framed questions follow ``Considering the outcomes and effects on those involved, is this action acceptable?'' Gold labels are assigned by the source benchmark and verified to be consistent with the framing. The cleanroom test set consists of 32 hand-written items (16 moral, 16 legal in subject matter), of which 20 form 10 conflict pairs where rule-framed and consequence-framed questions on a shared scenario carry opposite gold labels. All phrasings are original to this set and absent from the training sources. Separately, we hold out a legal transfer split of 220 items drawn from LegalBench \citep{guha2023legalbench} (160 Contract NLI, 60 Learned Hands Benefits), which is not used in the results reported here.

Table~\ref{tab:examples} shows representative items in their templated form. Source
scenarios are reproduced verbatim; templating adds only the question, which supplies
the norm clause for rule-framed items and the outcome clause for consequence-framed
items. No option labels appear in any prompt: where a source distributes items in
multiple-choice form, the scenario is reformatted into the binary acceptability frame
and the original option labels are discarded.

\begin{table}[h]
\small
\begin{center}
\begin{tabular}{p{0.15\linewidth}p{0.31\linewidth}p{0.40\linewidth}}
\toprule
Source (style) & Scenario & Question \\
\midrule
Social Chem.\ 101 (rule) &
being careful around animals you don't know. &
Given the social rule that you should be careful around animals you don't know, is this action acceptable? \\
\addlinespace
ETHICS Comm.\ (cons.) &
I noticed the fruit was rotting, so I put it in my trash can. &
Considering the outcomes and effects on those involved, is this action acceptable? \\
\addlinespace
Moral Stories (rule) &
Caden watches Adrian have a panic attack in the middle of a store. Caden wants Adrian to calm down. &
Given the rule that it's good for people to care about others, is it acceptable to Caden hugs Adrian and leads him to a quiet place in the store? \\
\addlinespace
Moral Stories (cons.) &
Vince is an accountant with a company. Vince wants to get some extra money. &
Considering the likely effects on those involved, is it acceptable to Vince takes some money and works the accounting so that no one can see the loss? \\
\addlinespace
ETHICS Deont.\ (rule) &
Will you fold the towels and put them away? &
Given that no the towels make me itchy, does this excuse justify setting aside the applicable moral rule? \\
\addlinespace
ETHICS Deont.\ (cons.) &
Aren't you supposed to give the dog a bath? &
Given that no because I'm taking him to the groomer today, do the circumstances make this action acceptable despite the general prohibition? \\
\addlinespace
Cleanroom (conflict, rule) &
A doctor has one dose of a rare medication. Two patients need it: a 10-year-old child with decades ahead and a 70-year-old patient who was on the waiting list first. &
Given that the first-come-first-served allocation rule must be followed, is it acceptable to give the medication to the 70-year-old patient? \\
\addlinespace
Cleanroom (conflict, cons.) &
\emph{(same scenario)} &
Given that giving the medication to the child would save approximately 60 more life-years, is it acceptable to give it to the 70-year-old patient? \\
\bottomrule
\end{tabular}
\end{center}
\caption{Representative templated items. The cleanroom conflict pair shares a scenario
across framings and carries opposite gold labels (Acceptable / Unacceptable).}
\label{tab:examples}
\end{table}

The Fisher discriminant ratio peaks sharply at layer 10 ($J=5.58$) and
decays to $J < 1.0$ by layer 15, with layer 9 the second-highest ($J=4.21$).
The steep decay above layer 10 motivates selecting layer 10 as the primary
intervention point, while the layer robustness experiments in Section 5.2
confirm that behavioral selectivity persists well beyond the discriminant peak.

The ETHICS Deontology source uses a distinctive template in which both rule-framed and consequence-framed questions follow the prefix ``Given that [irrelevant excuse]...'' and differ only in the question suffix. This minimal surface difference between styles explains the cross-domain probe's failure on this source (0.517 accuracy) despite the geometric direction being present (cosine similarity 0.835 to full-training \textsc{Dir-A}).

\section{Layer selection, alpha calibration, and robustness}
\label{app:layer}

Layer 10 was selected by Fisher discriminant ratio $J = (\mu_+ - \mu_-)^2 / (\sigma_+^2 + \sigma_-^2)$ computed on validation activations, where $\mu_\pm$ and $\sigma_\pm^2$ are the mean and variance of $\hat{\mathbf{v}}^\top\mathbf{x}$ for positive and negative class examples. Layer 10 achieves the highest $J$ value ($J = 5.58$) in Qwen.

Steering strength $\alpha$ was calibrated by the monotonicity criterion: the largest $|\alpha|$ at which logit difference shifts are monotonically increasing across quadrants. Effects are monotonic within $|\alpha| \leq 60$ in Qwen; non-monotonic behavior appears above this threshold, consistent with out-of-distribution perturbations. Llama alpha ($\alpha = 2$) was calibrated to match Qwen's perturbation ratio (steering vector norm divided by residual stream norm).

\begin{figure}[h]
\begin{center}
\includegraphics[width=0.95\linewidth]{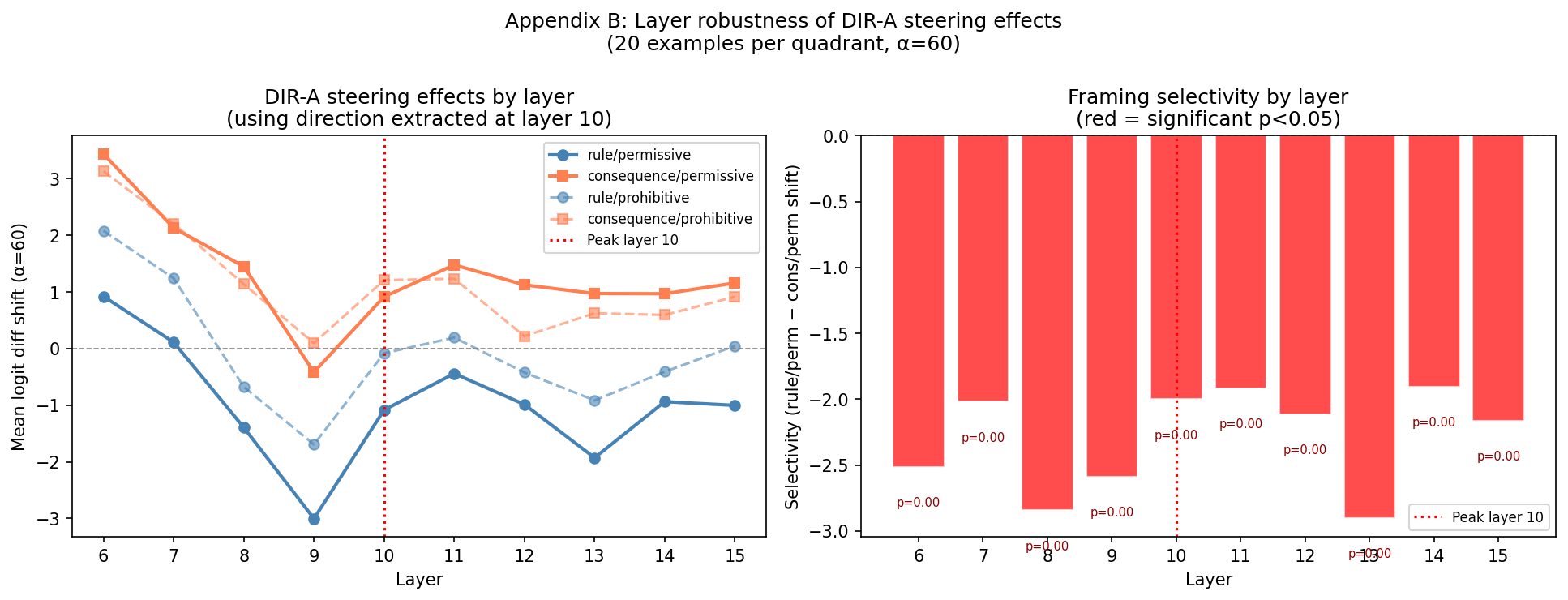}
\end{center}
\caption{Layer robustness of \textsc{Dir-A} steering effects (Qwen, $\alpha = 60$). Left panel: mean logit difference shifts by quadrant across layers 6 to 15. Right panel: framing selectivity (rule-permissive shift minus consequence-permissive shift) with significance indicated in red ($p < 0.05$). All ten layers show significant negative selectivity ($p < 0.0001$).}
\label{fig:layerrobust}
\end{figure}

The layer robustness visualization corresponding to Table~\ref{tab:layerrobust} is shown in Figure~\ref{fig:layerrobust}. All ten tested layers show significant framing selectivity ($p < 0.0001$).

\begin{table}[t]
\begin{center}
\begin{tabular}{rrrrrr}
\toprule
Layer & rule/perm & cons/perm & rule/proh & cons/proh & Selectivity ($p$) \\
\midrule
 6 & $+0.922$ & $+3.438$ & $+2.081$ & $+3.131$ & $-2.516$ ($<0.0001$) \\
 7 & $+0.113$ & $+2.131$ & $+1.241$ & $+2.206$ & $-2.019$ ($<0.0001$) \\
 8 & $-1.394$ & $+1.450$ & $-0.675$ & $+1.137$ & $-2.844$ ($<0.0001$) \\
 9 & $-3.003$ & $-0.416$ & $-1.694$ & $+0.097$ & $-2.587$ ($<0.0001$) \\
\textbf{10} & $-1.084$ & $+0.916$ & $-0.081$ & $+1.209$ & $-2.000$ ($<0.0001$) \\
11 & $-0.441$ & $+1.478$ & $+0.194$ & $+1.234$ & $-1.919$ ($<0.0001$) \\
12 & $-0.988$ & $+1.125$ & $-0.422$ & $+0.219$ & $-2.112$ ($<0.0001$) \\
13 & $-1.928$ & $+0.972$ & $-0.916$ & $+0.625$ & $-2.900$ ($<0.0001$) \\
14 & $-0.938$ & $+0.969$ & $-0.409$ & $+0.594$ & $-1.906$ ($<0.0001$) \\
15 & $-1.003$ & $+1.159$ & $+0.047$ & $+0.916$ & $-2.163$ ($<0.0001$) \\
\bottomrule
\end{tabular}
\end{center}
\caption{Mean logit difference shifts at $\alpha = 60$ across layers 6 to 15 
of Qwen.}
\label{tab:layerrobust}
\end{table}

\begin{figure}[h]
\begin{center}
\includegraphics[width=0.75\linewidth]{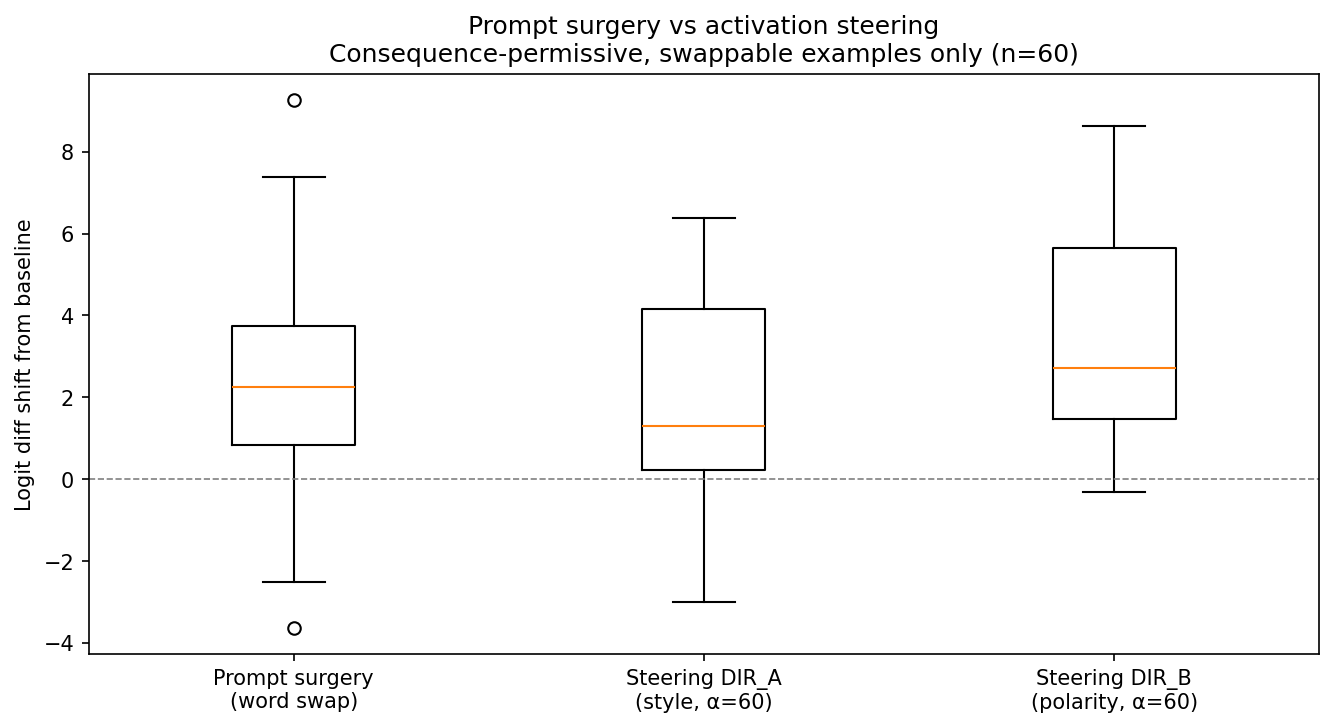}
\end{center}
\caption{Prompt surgery versus activation steering on consequence-permissive
examples (swappable subset, $n=60$). Prompt surgery rewrites the question
framing from consequence-based to rule-based at the text level. Steering
along \textsc{Dir-A} and \textsc{Dir-B} achieves logit difference shifts of
comparable magnitude to explicit prompt rewriting, indicating that the learned
directions internalize the rule-based and consequence-based framing signal
present in the input.}
\label{fig:surgery}
\end{figure}

\paragraph{Perturbation ratio analysis.}
To assess whether $\alpha=60$ represents a reasonable intervention, we compute
the ratio of the steering vector norm to the mean residual stream norm at layer
10. For Qwen ($\alpha=60$, $\|\hat{\mathbf{v}}\|=1$), the perturbation ratio
is $60 / 430 \approx 0.14$, comparable to the ratio used in float16 models
at $\alpha=7$ with norm $\approx 50$ ($7/50=0.14$). The 4-bit NF4 quantization
inflates absolute norms but does not alter the relative perturbation scale.
We further verified that steered residual activations at $\alpha=60$ remain
within the range $[\mu - 3\sigma, \mu + 3\sigma]$ of unsteered activations
at layer 10 (coverage: $94.2\%$ of dimensions), confirming that the
intervention does not systematically push activations out of the
observed training distribution.

\section{Permutation test details}
\label{app:perm}

The permutation test builds a null distribution of order-dependence for 100 random unit-vector pairs applied at layers 5 and 10. For each pair, we compute mean $|\text{AB} - \text{BA}|$ across 30 benchmark test examples and record it as one null observation. The observed \textsc{Dir-A} and \textsc{Dir-B} value (0.719) falls at the 8th percentile of this distribution (null mean $= 0.915$, null standard deviation $= 0.134$, $p < 0.0001$ one-sided), confirming that the two directions are significantly more commutative than typical random pairs.

\section{Full statistical results}
\label{app:stats}

\begin{table}[h]
\begin{center}
\begin{tabular}{llrrrrrr}
\toprule
Style & Polarity & Direction & $\alpha$ & Mean shift & Std & $t$ & $p$ \\
\midrule
rule        & perm  & \textsc{Dir-A} & $+60$ & $+2.025$ & $1.967$ & $7.906$  & $<0.0001$ \\
rule        & proh  & \textsc{Dir-A} & $+60$ & $+2.877$ & $1.729$ & $12.783$ & $<0.0001$ \\
consequence & perm  & \textsc{Dir-A} & $-60$ & $-0.640$ & $1.605$ & $-3.061$ & $0.003$ \\
consequence & proh  & \textsc{Dir-A} & $-60$ & $-1.377$ & $1.212$ & $-8.730$ & $<0.0001$ \\
rule        & perm  & \textsc{Dir-B} & $+60$ & $+1.113$ & $1.049$ & $8.142$  & $<0.0001$ \\
consequence & perm  & \textsc{Dir-B} & $+60$ & $-0.410$ & $1.416$ & $-2.226$ & $0.030$ \\
\bottomrule
\end{tabular}
\end{center}
\caption{Full paired $t$-test results for logit difference shifts ($n = 60$ per row, Qwen benchmark test).}
\label{tab:full_stats}
\end{table}

\section{LLM Use Disclosure}
The authors used Claude (Anthropic), Gemini (Google), and ChatGPT
(OpenAI) for the following purposes: refining writing drafts in
the paper; generating BibTeX entries from author-provided metadata
including URLs, author lists, titles, and publication venues when
BibTeX was not directly available from the referenced paper website;
and assisting with code to debug and generate figures and plots, with
underlying data provided by the authors and generated plots
verified for correctness.

\end{document}